\documentclass[conference]{IEEEtran}
\IEEEoverridecommandlockouts
\usepackage{cite}
\usepackage{amsmath,amssymb,amsfonts}
\usepackage{algorithmic}
\usepackage{graphicx}
\usepackage{textcomp}
\usepackage{xcolor}
\usepackage{orcidlink}
\def\BibTeX{{\rm B\kern-.05em{\sc i\kern-.025em b}\kern-.08em
    T\kern-.1667em\lower.7ex\hbox{E}\kern-.125emX}}

\usepackage[export]{adjustbox} % valign
\usepackage{array,ragged2e}
\usepackage{amsfonts}

\usepackage{caption}

\usepackage{makecell}

\usepackage{dsfont}
\usepackage{graphicx}
\usepackage{multirow}

\begin{document}
\bstctlcite{IEEEexample:BSTcontrol}

\title{Exploiting Multiple Sequence Lengths in Fast End to End Training for Image Captioning
\thanks{This work has received funding from the European
Union's Horizon 2020 programme dAIEDGE (G.A. No 101120726).}
}

% \author{
%     \IEEEauthorblockN{Jia Cheng Hu\IEEEauthorrefmark{1}, Roberto Cavicchioli\IEEEauthorrefmark{2}, Alessandro Capotondi \IEEEauthorrefmark{3}}
%     \IEEEauthorblockA{
%     \IEEEauthorrefmark{1}\IEEEauthorrefmark{3}\textit{Department of Physics, Informatics and Mathematics}\\
%     \IEEEauthorrefmark{2}\textit{Department of Communication and Economics}\\
%     \textit{University of Modena and Reggio Emilia, Italy}\\
%     \{name.surname\}@unimore.it\\
%     orcids: \IEEEauthorrefmark{1}0009-0008-1611-966X, \IEEEauthorrefmark{2}0000-0003-0166-0898, \IEEEauthorrefmark{3}0000-0001-8705-0761}
% }

\author{\IEEEauthorblockN{Jia Cheng Hu \orcidlink{0009-0008-1611-966X}}
\IEEEauthorblockA{\textit{FIM Dept.} \\
\textit{Univ. of Modena and Reggio Emilia}\\
Modena, Italy \\
jiacheng.hu@unimore.it}

\and
\IEEEauthorblockN{Roberto Cavicchioli \orcidlink{orcid=0000-0003-0166-0898}}
\IEEEauthorblockA{\textit{DCE Dept.} \\
\textit{Univ. of Modena and Reggio Emilia}\\
Reggio Emilia, Italy \\
roberto.cavicchioli@unimore.it}
\and
\IEEEauthorblockN{Alessandro Capotondi \orcidlink{0000-0001-8705-0761}}
\IEEEauthorblockA{\textit{FIM Dept.} \\
\textit{Univ. of Modena and Reggio Emilia}\\
Modena, Italy \\
alessandro.capotondi@unimore.it}
}

\maketitle

\begin{abstract}
We introduce a method called the Expansion mechanism that processes the input unconstrained by the number of elements in the sequence. By doing so, the model can learn more effectively compared to traditional attention-based approaches. To support this claim, we design a novel architecture ExpansionNet v2 that achieved strong results on the MS COCO 2014 Image Captioning challenge and the State of the Art in its respective category, with a score of 143.7 CIDErD in the offline test split, 140.8 CIDErD in the online evaluation server and 72.9 AllCIDEr on the nocaps validation set. Additionally, we introduce an End to End training algorithm up to 2.8 times faster than established alternatives.

% *CRITICAL: Do Not Use Symbols, Special Characters, Footnotes, or Math in Paper Title or Abstract. -> questo mi ha fatto cambiare il titolo 
\end{abstract}

\section{Introduction}

Image Captioning consists of the problem of describing images without human intervention. It is a challenging multi-modal task that requires both language comprehension and visual understanding. Early approaches relied on statistical and graph-based methods \cite{mitchell2012midge, kulkarni2013babytalk}, but since the advent of Neural Networks most Image Captioning systems adopted an encoder and decoder structure \cite{vinyals2015show, xu2015show, anderson2018bottom}. The first component is responsible for extracting visual features from the image, whereas the latter serves the purpose of generating the description. Early works \cite{xu2015show, vinyals2015show, anderson2018bottom} relied on Convolutional Neural Network (CNN) backbones \cite{ren2015faster} combined with Recurrent Neural Networks (RNNs) \cite{hochreiter1997long, cho2014learning} to further refine the visual inputs and for text generation. In contrast, modern Image Captioning systems adopt Attention-based \cite{bahdanau2014neural, vaswani2017attention} architectures  for the sequence modelling part and, in recent works \cite{wang2022end, nguyen2022grit, liu2021swin}, also during the image feature extraction. Currently, fully attentive models are the standard de facto architecture in many NLP and Vision research fields and their ubiquity led to many refinements and improvements of the formulation across multiple fields \cite{pan2020x, nguyen2022grit, wang2022ofa, wang2022git, herdade2019image, luo2021dual, wang2022end, guo2019star, hao2019modeling}. However, one of the purposes of the development Attention mechanism \cite{bahdanau2014neural, luong2015effective} was to spread the input sequence content along the whole collection of encoder's hidden vectors instead of one single state, overcoming a significant performance bottleneck in RNNs. To do so, as the name suggests, the Attention mechanism enhances the values of a few elements and inhibits the others by
\begin{figure}[htbp]
\centering
\includegraphics[width=0.7\columnwidth]{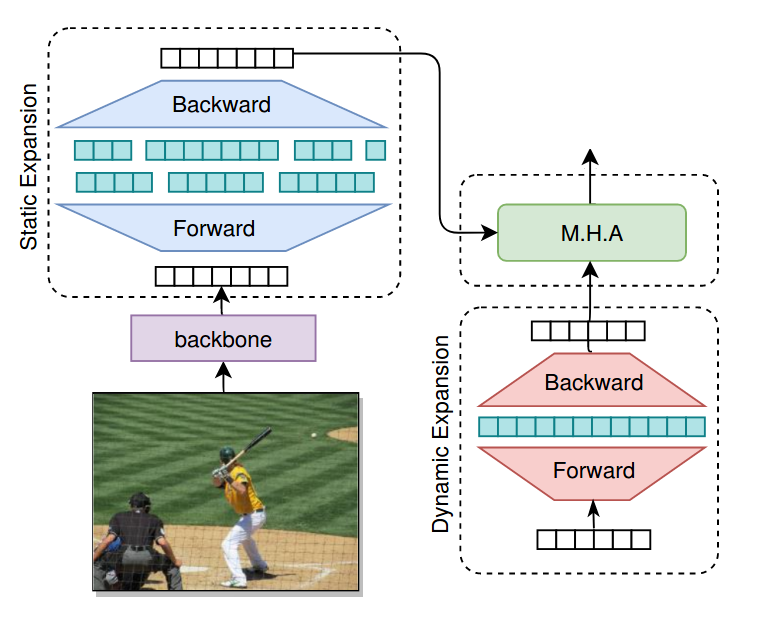}
\caption{The expansion mechanism distributes the input data into another one featuring a different sequence length during the forward phase and performs the reverse operation in the backward pass. In this way, the network is enabled to process the sequence unconstrained by the number of elements.}
\label{fig:expansion_mechanism_intro}  
\end{figure}
means of the Softmax function. Recently, many studies  \cite{raganato2020fixed, tay2021synthesizer, ramsauer2020hopfield, you2020hard, lee2021fnet, tolstikhin2021mlp} deepened the 
understanding of attention approach and suggested that there is little difference between the first and alternative solutions such as Gaussian distributions \cite{you2020hard}, MLPs \cite{tolstikhin2021mlp}, Fourier Transform \cite{lee2021fnet} and suggested that the effectiveness of these methods depends mainly on their capability to form high-quality compositions out of the input. % ecco il problema di ricerca... investighiamo se per caso la lunghezza fissa di sequenza rappresenta un ostacolo alla formazione di compositions migliori
Motivated by these observations, our work investigates the possibility that the fixed number of elements provided by the input (the sequence length) represents a performance bottleneck for stateless architectures and limits their potential to form higher-quality compositions, in the particular field of Image Captioning. 
% c'e' un sacco di ripetizione qui sopra... da sistemare.
To this end, we propose the Expansion mechanism, a method that distributes and processes the sequence content using an increased or arbitrary number of elements and retrieves the original length back in the complementary backward operation. We then introduce ExpansionNet v2 (depicted in Fig. \ref{fig:expansion_mechanism_intro}), which to our knowledge is the first model that learns to exploit arbitrary sequence lengths in Image Captioning and achieves very competitive results without relying on the Attention's characteristic function.

The overall contributions of this work are the following: (i) we introduce a new method called Expansion Mechanism that distributes the input content over an arbitrary or increased number of 
elements during the forward step, and retrieves the original length back in the complementary backward operation. To support both bidirectional and auto-regressive processing, we introduce two methods, called Static Expansion and Dynamic Expansion. The efficiency aspect
is addressed in their design and as a result, the computational impact is negligible for small configurations; (ii) with the aforementioned methods, we design a novel architecture called ExpansionNet v2 that achieves strong results on the MS-COCO 2014 outperforming similar models trained on the same dataset; (iii) given the positive results of our architecture, we find out that traditional architectures in Image Captioning are indeed penalized by the fixed number of elements provided by the input; 
(iv) in contrast to the general trend, our achieves strong results despite the removal of the Attention in most components. Finally, we also propose a fast End-to-End training strategy that lowers significantly the training cost of our model compared to popular approaches.

\section{Related Works}

Image Captioning models benefited greatly from Deep Learning methods. From hand-crafted sentences combined with object detection \cite{socher2010connecting, yao2010i2t}, modern systems consist of a neural encoder that extracts meaningful visual representations from the image and a decoder responsible for the description generation. 
In the early formulations, the decoder consisted of RNNs \cite{cho2014learning, hochreiter1997long}, whereas the encoder consisted of a convolutional backbone \cite{vinyals2015show, xu2015show} that represented the entire image with a single feature vector. It was later replaced by an object detector \cite{anderson2018bottom} that extracted a collection of salient regions of the image. This enabled the adoption of sequence modelling architectures in both encoding and decoding \cite{wang2020show, anderson2018bottom, xu2015show, vinyals2015show} on top of the backbones. Most modern Image Captioning systems are currently based on the Transformer architecture \cite{vaswani2017attention} and many works focused on improving its formulation or structure \cite{huang2019attention, herdade2019image, kim2018bilinear, pan2020x, guo2019star, sukhbaatar2019augmenting, gulati2020conformer}. For example, the work of \cite{herdade2019image} introduced geometrical awareness in the Self-Attention formulation. \cite{huang2019attention} modified the attentive layer with a gate that served the purpose of mitigating the contribution of irrelevant queries. \cite{pan2020x} exploited the bilinear pooling to enable a higher order of interactions across the input elements. 
Other works such as \cite{luo2021dual, ji2021improving, nguyen2022grit} focused on structural changes and exploiting the visual input more effectively. Overall, all these methods follow the main components of the formulas introduced in \cite{bahdanau2014neural, luong2015effective, vaswani2017attention}. Our Expansion mechanism is based on the adoption of embedding vectors. The effectiveness of integrating additional learnable parameters in the sequence was observed first in \cite{sukhbaatar2019augmenting} in Machine Translation. Later in Image Captioning, the concept was also deployed by \cite{cornia2020meshed} and \cite{zeng2022s2}. In contrast to these works, our method is the only one that distributes the input into an arbitrary number of hidden vectors.

Another trend consists of pre-training the model with a huge amount of training data and fine-tuning over the Image Captioning task \cite{wang2022unifying, wang2022git, hu2022scaling, li2022blip}.  
In particular, OFA \cite{wang2022unifying} and GIT \cite{wang2022git} currently represent the State-of-the-Art Image Captioning systems and outperform non-generative models by a significant margin. However, their model size poses an obstacle to the deployment in memory-limited devices and the training data are tens and hundreds of times bigger than the popular MS-COCO 2014 \cite{lin2014microsoft}. For this reason, these works are considered orthogonal to ours which can be instead integrated to potentially achieve better performances. In general, we only consider works that are trained exclusively on MS-COCO 2014, for this reason, the works of \cite{nguyen2022grit, wang2022ofa, wang2022git, hu2022scaling, li2022blip} are omitted during evaluation since our model does not leverage additional data.

% Figura schematica
\begin{figure*}[htbp]
\centering
\includegraphics[width=1.0\textwidth]{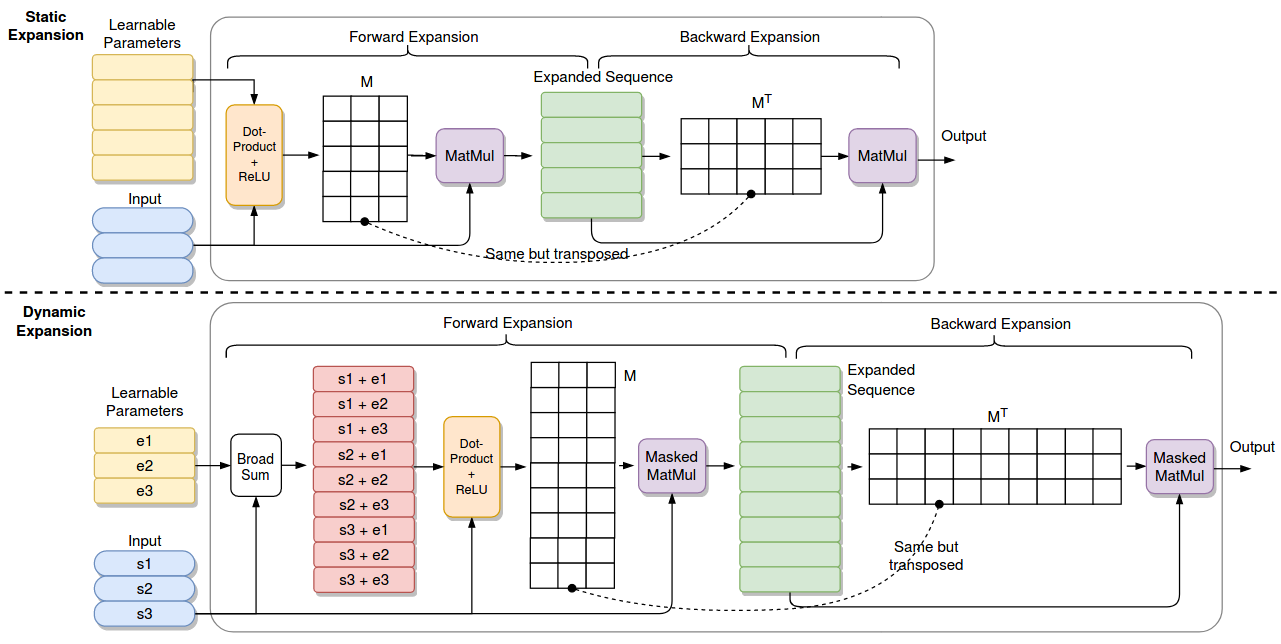}
\caption{Static Expansion and Auto-regressive Dynamic Expansion scheme and example. Assuming an input length of $L=3$. In the Static Expansion setting, an expansion coefficient of $N_{E}=5$ leads to an expanded sequence of length $5$. In contrast, in the Dynamic Expansion, an expansion coefficient of $N_{E}=3$ generates an expanded sequence of $L \cdot N_{E} = 9$. For the sake of simplicity, the double operation stream, the expansion biases and the gated result combination are omitted in the illustration. The difference between the Auto-regressive Dynamic Expansion and the bidirectional one lies in the Masked Matrix Multiplication. }
\label{fig:forward_backward_scheme.png}
\end{figure*}

\section{Method}

\subsection{Static and Dynamic Expansion}

The Expansion mechanism is broken down into several steps. First, it distributes the sequence content into an arbitrary or increased number of elements (Section \ref{section:expansion_coefficient}) using a ``Forward Expansion'', which is described in Section \ref{section:forward_expansion} and allows the network to process the sequence unconstrained by the fixed input length. Then, it retrieves the original length using the complementary operation ``Backward Expansion'', described in Section \ref{section:backward_expansion}. Depending on the operations, we define two implementations of the idea: Static Expansion and Dynamic Expansion. The latter is designed to support both the auto-regressive and bidirectional processing, in contrast to the first, which only supports the bidirectional case. 

\subsubsection{Expansion coefficient}
\label{section:expansion_coefficient}
In both Static and Dynamic Expansion, the expansion coefficient $N_{E}$ defines a collection of learnable parameters $E_{Q}, E_{B} \in \mathbb{R}^{N_{E} \times d_{m}}$.
However, in the Static Expansion, $N_{E}$ defines exactly the size of the expanded sequences regardless of the input length $L$. In particular, the expansion queries $Q_{E}$ and biases $B_{E}$ equal to $E_{Q}$ and $E_{B}$ respectively. In contrast, in the Dynamic Expansion, the expanded sequence is of size $N_{E} \cdot L$, and the expansion queries $Q^{E}$ and biases $B_{E}$ are calculated with the BroadSum operator, defined in the two cases as:
\begin{equation}
    \begin{aligned}
        Q_{E} &= (C^{\top}\mathbb{H}_{E})^{\top} + (E_{Q}^{\top}\mathbb{I}_{E})^{\top} \\ 
        B_{E} &= (C^{\top}\mathbb{H}_{E})^{\top} + (E_{B}^{\top}\mathbb{I}_{E})^{\top}
    \end{aligned}
\end{equation}
where $C \in \mathbb{R}^{L \times d_m}$ denotes a linear projection of the input and $\mathbb{H}_{E} \in \mathbb{R}^{L \times (L \cdot N_{E})}$ is defined as:
\[
\mathbb{H}_{E}
=
\begin{bmatrix}
\mathds{1} & \mathbf{0} & \dots & \mathbf{0} \\
\mathbf{0} & \mathds{1} & \dots & \mathbf{0} \\
\vdots & \vdots & \ddots & \vdots \\
\mathbf{0} & \mathbf{0} & \dots & \mathds{1} \\
\end{bmatrix}
, 
\ \ \mathds{1}, \mathbf{0} \in \mathbb{R}^{1 \times N_{E}}
\]
whereas $\mathbb{I}_{E} \in \mathbb{R}^{N_{E} \times (L \cdot N_{E})}$ is defined by the column-wise concatenation of $L$ identity matrices of size $N_{E} \times N_{E}$:
\[
\mathbb{I}_{E}
=
\begin{bmatrix}
\textnormal{I}_{L} & \textnormal{I}_{L} & \dots & \textnormal{I}_{L} \\
\end{bmatrix}
, 
\ \ \textnormal{I}_{L} \in \mathbb{R}^{N_{E} \times N_{E}}
\].
An example of the input and output of the BroadSum operation is depicted in the bottom left of Figure \ref{fig:forward_backward_scheme.png}, where the bias vectors are omitted for simplicity.

\subsubsection{Forward Expansion}
\label{section:forward_expansion}
The forward expansion generates the expanded sequences and involves three linear projections of the input, denoted as $K, V_{1}, V_{2} \in \mathbb{R}^{L \times d_{m}}$. First of all, the ``Length Transformation Matrix'', denoted as $M$, is computed as the dot-product similarity between $K$ and the expansion queries $Q_{E}$:
\begin{equation}
    M =  \frac{Q_{E}K^\top}{\sqrt{d_{m}}}.
\label{eq:sim_forward}
\end{equation}
The result is fed into the following operations:
\begin{equation}
\begin{aligned}
R^{fw}_{i} &= \Psi(ReLU((-1)^{i}M), \epsilon) \ \ \ \ i \in \{1, 2\} \\
\end{aligned}
\label{eq:forward_matrix}
\end{equation}
where $\Psi: (X, \epsilon) \rightarrow Y$, $X,Y \in \mathbb{R}^{N_{1} \times N_{2}}, \epsilon \in \mathbb{R}^{+}/\{0\}$ is the row-wise normalization function defined as:
\begin{equation}
    \Psi(X, \epsilon)_{ij} = \frac{x_{ij}}{\sum_{z=1}^{N_{2}}x_{iz} + \epsilon} 
    \label{eq:norm}
\end{equation}
the coefficient $\epsilon$ ensures the feasibility of the operation. Then, the expanded sequences are calculated as follows:
\begin{equation}
\begin{aligned}
F^{fw}_{i} &= R^{fw}_{i}V_{i} + B_{E} \ \ \ \ i \in \{1, 2\} \\
\end{aligned}
\label{eq:forward_results}
\end{equation}

\subsubsection{Backward expansion}
\label{section:backward_expansion}

In the backward step, the original sequence length is retrieved by transposing the length transformation matrix in Equation \ref{eq:sim_forward} and applying the same operations of Equation \ref{eq:forward_matrix}:
\begin{equation}
\begin{aligned}
R^{bw}_{i} &= \Psi(ReLU((-1)^{i}M^\top), \epsilon) \ \ \ \ i \in \{1, 2\} \\
\end{aligned}
\label{eq:backward_computation}
\end{equation}
This time, the matrices $R^{bw}_{i}$ are multiplied with the expanded sequences of Equation \ref{eq:forward_results}:
\begin{equation}
\begin{aligned}
B^{bw}_{i} &= R^{bw}_{i}F^{fw}_{i}  \ \ \ \ i \in \{1, 2\} \\
\end{aligned}
\label{eq:backward_results}
\end{equation}
Finally, the final results $B^{bw}_{1}$ and $B^{bw}_{2}$ are combined by means of a sigmoid gate:
\begin{equation}
out = \sigma(S) \odot B^{bw}_{1} + (1 - \sigma(S))
\odot B^{bw}_{2}.
\end{equation}
where $S \in \mathbb{R}^{L}$ is a linear projection of the input.

% immagine dell'architettura
\begin{figure*}[htbp]
\centering
\includegraphics[width=0.65\textwidth]{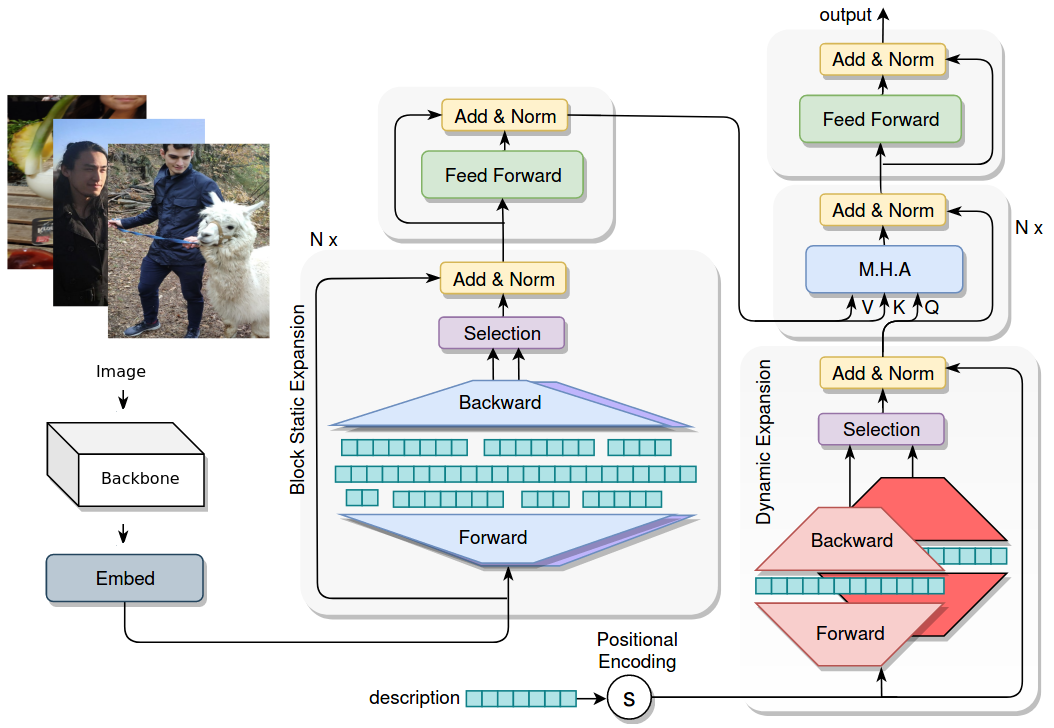}
\caption{ExpansionNet v2 architecture.}
\label{fig:intera_architettura_showcase.png}
\end{figure*}

The backward operation completes the operations performed in the Static and Dynamic Expansion. It can be noted that all operations in the forward (\ref{eq:forward_matrix})(\ref{eq:forward_results}) and backward expansion (\ref{eq:backward_results}) (\ref{eq:backward_computation}) are duplicated in two operations streams for $i=1$ and $i=2$, differing mainly in the sign of the computation of the Length Transformation Matrix in (\ref{eq:sim_forward}). The decision was made to mitigate the remote possibility of the matrix being populated only by zeros. This does not affect the results compared to the single path but slightly increases the computational cost.

In the case of Dynamic Expansion, masking is applied when calculating the results in (\ref{eq:forward_results}) and (\ref{eq:backward_results}) to preserve the auto-regressive property. The operation principle of Static and Dynamic Expansions are illustrated in Fig. \ref{fig:forward_backward_scheme.png}, which, for simplicity, depicts only a single operation stream and omits biases and the output sigmoid gates.

\subsubsection{Block Static Expansion}
To increase the effectiveness of the Static Expansion, we perform the Forward and Backward operations on a collection of target lengths instead of one. We call the operation Block Static Expansion.
From a formulation perspective, all operations are repeated over a group of expansion coefficients $G = \{N^1_E, N^2_E, \ldots, N^{N_G}_E\}$ and can be implemented in a way such that both forward and backward steps are performed over all targets at the same time. All expansion group queries and biases can be combined into a single one:
\begin{equation}
\begin{aligned}
    E^{G}_{Q} &= \{(E^{1}_{Q})^\top, (E^{2}_{Q})^\top, \ldots, (E^{N_{G}}_{Q})^\top\}^\top \\
    E^{G}_{B} &= \{ (E^{1}_{B})^\top, (E^{2}_{B})^\top, \ldots, (E^{N_{G}}_{B})^\top \}^\top
\end{aligned}
\label{equation_block_static_generation}
\end{equation}
and the computational efficiency of the previous formulation can be preserved.
During the backward stage, the length transformation matrix is scaled by the inverse number of elements in the group $G$.

\subsection {Architecture}
Our model consists of the standard encoder-decoder structure implemented on top of the Swin-Transformer, which details are provided in \cite{liu2021swin}. The image $A$ is first fed into the backbone:
\begin{equation}
    \begin{aligned}
        X_{0} &= Swin\textnormal{-}Transf(A) 
    \end{aligned}
\end{equation}
and generates the initial set of processed visual features  $X_{0}=\{x^{0}_1, x^{0}_2, \ldots, x^{0}_N\}, \ x^{0}_{i} \in \mathbb{R}^{d_m}$. The result is fed into the encoder, which is made of $N_{enc}$ Static Expansion $\rightarrow$ FeedForward blocks. Here skip connection and pre-layer normalization \cite{xiong2020layer} are adopted, and the following formulas describe each encoder layer for $n \in \{1,\ldots, N_{enc}\}$:
\begin{equation}
    \begin{aligned}
        E_{n} &= X_{n-1} + StaticExp_{n}(Norm^{SE}_{n}(X_{n-1})) \\
        X_{n} &= E_{n} + FF_{n}(Norm^{FF}_{n}(B_{n}))
    \end{aligned}
\end{equation}
Similarly, given a generic input sequence $Y_{0}=\{y^{0}_1, y^{0}_2, \ldots, y^{0}_M\}, \ y^{0}_{i} \in \mathbb{R}^{d_m}$ (at training stage so we can omit the time axis), the decoder is made of $N_{dec}$ Dynamic Expansion $\rightarrow$ Cross-Attention $\rightarrow$ FeedForward blocks, where skip connection and normalization is applied on each component. Each decoder layer is described by the following equations:
\begin{equation}
    \begin{aligned}
        B_{n} &= Y_{n-1} + DynamicExp_{n}(Norm^{DE}_{n}(Y_{n-1})) \\
        W_{n} &= B_{n} + Attention_{n}(Norm^{CA}_{n}(B_{n}), X_{N_{enc}}) \\
        Y_{n} &= W_{n} + FF_{n}(Norm^{FF}_{n}(W_{n}))
    \end{aligned}
\end{equation}
All layers are summed through a linear projection and the final output is fed to the classification layer. 
Fig. \ref{fig:intera_architettura_showcase.png} depicts the main structure.

\begin{table*}[htbp]
  \begin{center}
  \caption{Ablation study in the first stage of Cross-Entropy training using beam size 3 over the Karpathy validation split. B=BLEU. M=METEOR. R=ROUGE. C=CIDEr-D. S=SPICE.}
  \begin{tabular}{| c | c | c | c | c | c | c | c | c | c |} 
 \hline
 \textbf{Encoder} & \textbf{Decoder} & \textbf{B1} & \textbf{B2} & \textbf{B3} & \textbf{B4} & \textbf{M} & \textbf{R} & \textbf{C} & \textbf{S} \\
 \hline
 Baseline & Baseline 
 & 75.3 & 59.2 & 45.4 & 34.6 & 28.4 & 57.0 & 115.8 & 21.6 \\
 Stc. Exp. G=$\{ 16 \}$ & Baseline & 76.4 & 60.6 & 46.6 & 35.5 & 28.6 & 57.2 & 117.8 & 21.9  \\
 Stc. Exp. G=$\{ 32 \}$ & Baseline & 75.9 & 59.9 & 46.1 & 35.2 & 28.9 & 57.1 & 117.9 & 22.3 \\
 Stc. Exp. G=$\{ 64 \}$ & Baseline & 76.3 & 60.4 & 46.4 & 35.5 & 28.8 & 57.1 & 117.7 & 22.0 \\
Baseline & Dyn. Exp.\textsubscript{$N_E$=4} & 77.2 & 61.4 & 47.4 & 36.2 & 28.9 & 57.7 & 119.7 & 22.3 \\

Baseline & Dyn. Exp.\textsubscript{$N_E$=8} & 76.9 & 61.5 & 47.9 & 37.1 & 29.1 & 57.8 & 120.8 & 22.3 \\
               
Baseline & Dyn. Exp.\textsubscript{$N_E$=16} & 76.7 & 61.4 & 47.8 & 36.8 & 29.0 & 57.7 & 121.2 & 22.2 \\

Stc. Exp. G=$\{ 64 \}$ & Dyn. Exp.\textsubscript{$N_E$=16} 
 & 77.4 & 61.9 & 48.2 & 37.3 & 29.2 & 58.0 & 122.2 & 22.3 \\
                   
 Stc. Exp. G=$\{ 128, 128, 128, 128, 128 \}$ & Dyn. Exp.\textsubscript{$N_E$=16} 
 & 77.8 & 62.3 & 48.3 & 37.2 & 29.3 & 58.3 & 122.8 & 22.5 \\                                          
 Stc. Exp. G=$\{ 256, 256, 256, 256, 256 \}$ & Dyn. Exp.\textsubscript{$N_E$=16} 
 & 77.4 & 62.0 & 48.2 & 37.2 & 29.2 & 58.0 & 122.5 & 22.2 \\   
 Stc. Exp. G=$\{ 512, 512, 512, 512, 512 \}$ & Dyn. Exp.\textsubscript{$N_E$=16} 
 & 77.3 & 61.7 & 47.9 & 37.0 & 29.3 & 58.0 & 122.7 & 22.4 \\
 Stc. Exp. G=$\{ 32, 64, 128, 256, 512 \}$ & Dyn. Exp.\textsubscript{$N_E$=16} & 77.6 & 62.0 & 48.2 & 37.2 & 29.4 & 58.1 &  123.5 & 22.5  \\
 \hline
 \end{tabular}
 \label{tab:ablation_1}
\end{center}
\end{table*}

\subsection{Training objectives}

The model is first pre-trained using the Cross-Entropy loss $L_{XE}$:
\begin{equation}
    L_{XE}(\theta) = - \sum_{t}^{T} log( p_\theta(y^{*}_{t}|y^{*}_{1:t-1}, I))
\end{equation}
where $p_\theta(y^{*}_{t}|y^{*}_{1:t-1}, I)$ is the  probability assigned by the model parameters $\theta$ to the target $y^{*}_{t}$ given the image $I$ and the previous words $y^{*}_{1:t-1}$. Additionally, the CIDEr-D score is optimized using the SCST \cite{rennie2017self} which minimizes the negative expected reward $L_{R}(\theta)=-\mathbb{E}_{y_{1:T~p_{\theta}}}[r(y_{1:T})]$, which gradient can be approximated as follows:
\begin{equation}
        \nabla_{\theta} L_{R}(\theta) \approx -(r(y_{1:T}^s) - b) \nabla_\theta log \  p_\theta(y_{1:T}^s)
    \label{scst_loss}
    %L_{RF}(\theta) = -(r - b) \cdot \sum_{t}^{M} log( p_\theta(y_{t}|y_{1:t-1}))
\end{equation}
$b$ is the baseline computed according to \cite{luo2020better} and $r(y_{1:T}^{s})$ is the CIDEr-D reward assigned to the sampled sequence $y_{1:T}^{s}$. 

Although we optimize the model on two loss functions, for each one of them, the training stage is efficiently split into two additional steps to allow a broader number of computational resources to reproduce this work.

\section{Results}

\subsection{Experimental Setup}
\label{training_section}

\subsubsection{Dataset} The training dataset consists of the popular MS-COCO benchmark \cite{lin2014microsoft} split according to \cite{karpathy2015deep}, resulting in 113287 image-description pairs for the training, 5000 in the validation set, and in the 5000 test set. Each reference caption is pre-processed by a simple pipeline consisting of lowering casing, removing punctuation, and filtering out words that do not occur at least 5 times (vocabulary of size 10000). Additionally, the final model is evaluated over the Novel Object Captioning at Scale (nocaps) dataset validation set \cite{agrawal2019nocaps}, which consists of three classes of images called in-domain, near-domain, and out-domain, according to the familiarity of the classes with respect to those contained in the training set. This dataset is subject to the same pre-processing of MS-COCO and serves the purpose of further challenging the model in unfavourable conditions.

\subsubsection{Model details} Two models are implemented for the experimental setup. The baseline, which is the Base Transformer and our main model, referred to as ``ExpansionNet v2", is implemented with the following configurations $d_{m}$=$512$, $d_{ff}$=$2048$, $N_{enc}$=$N_{dec}$=$3$. In the latter, the Dynamic expansion coefficient is set to 16, and the Static expansion coefficients consist of $G$=$\{32, 64, 128, 256, 512\}$ (more details in Section \ref{section_ablation}). Each one relies on top of the same backbone, the Swin-Transformer in the Large configuration \cite{liu2021swin} pre-trained on ImageNet \cite{deng2009imagenet}. All images are subject to a minimal pre-processing: first, they are resized into a $3{\times}384{\times}384$ tensor, then RGB values are converted into a $[0, 1]$ range and further normalized using $mean$=$(0.485,0.456,0.406)$ and $std$=$(0.229,0.224,0.225)$. The source code of the experiments is available\footnote{Code available at: https://github.com/jchenghu/ExpansionNet\_v2}.

\subsubsection{Training algorithm} It can be observed that the Swin-Transformer backbone is the most computationally expensive part of the system.
For this reason, inspired by \cite{hu2022exploring}, to enable the End to End training step to a broader number of computational architectures, our training is divided into four steps in particular, each phase (in both the cross-entropy training and the reinforcement stage) consists of initial training in which the backbone's weights are frozen and a fine-tuning step during which gradients flow throughout the whole system:
\par \textit{Step A) Cross-Entropy -- Freezed backbone.} The model is trained using batch size 48, an initial learning rate of 2e-4, a warmup of 10000, and is annealed by 0.8 every 2 epochs for 8 epochs;
\par \textit{Step B) Cross Entropy -- End to End.} The whole system is trained for 2 additional epochs, using batch size 48 and an initial learning rate of 3e-5 annealed by 0.55 every epoch;
\par \textit{Step C) CIDEr-D optimization -- Freezed backbone.} Reinforcement phase adopts a batch size of 48, an initial learning rate of 1e-4, no warmup, annealed by 0.8 every epoch for 9 epochs;
\par \textit{Step D) CIDEr-D optimization -- End to End.} The whole system is fine-tuned for a few more iterations up to an additional epoch using a batch size of 20 and fixed learning rate 2e-6. This step is optional since it only slightly contributes to the final performances and can be skipped if no improvements are observed. All CIDEr-D optimization steps are implemented according to the Standard configuration\footnote{SacreEOS signature \cite{Hu202339}: \\ STANDARD\_wInit+Cider-D[n4,s6.0]+average[nspi5]+1.0.0}.

% offline scores

\begin{table*}[htbp]
  \begin{center}
  \caption{Offline comparison of State-of-the-Art single models over the Karpathy test split. B=BLEU. M=METEOR. R=ROUGE. C=CIDEr-D. S=SPICE.}
\label{tab:offline_table_eval}
  \begin{tabular}{| c | c  c  c  c  c  c | c  c  c  c  c  c |}
 \hline 
 {} & \multicolumn{6}{|c|}{\textbf{Cross-Entropy}} & \multicolumn{6}{c|}{\textbf{CIDEr-D optimization}} \\
 \cline{2-13}
 \textbf{Model} & \textbf{B1} & \textbf{B4} & \textbf{M} & \textbf{R} & \textbf{C} & \textbf{S} & \textbf{B1} & \textbf{B4} & \textbf{M} & \textbf{R} & \textbf{C} & \textbf{S} \\
 \hline
 Up-Down \cite{anderson2018bottom} & 77.2 & 36.2 & 27.0 & 56.4 & 113.5 & 20.3 & 79.8 & 36.3 & 27.7 & 56.9 & 120.1 & 21.4\\
 GCN-LSTM \cite{yao2018exploring} & 77.3 & 36.8 & 27.9 & 57.0 & 116.3 & 20.9 & 80.5 & 38.2 & 28.5 & 58.3 & 127.6 & 22.0\\
 SGAE \cite{yang2019auto} & - & - & - & - & - & - & 80.8 & 38.4 & 28.4 & 58.6 & 127.8 & 22.1\\
 AoANet \cite{huang2019attention}& 77.4 & 37.2 & 28.4 & 57.5 & 119.8 & 21.3 & 80.2 & 38.9 & 29.2 & 58.8 & 129.8 & 22.4\\
 X-Transformer \cite{pan2020x} & 77.3 & 37.0 & 28.7 & 57.5 & 120.0 & 21.8 & 80.9 & 39.7 & 29.5 & 59.1 & 132.8 & 23.4 \\
 GET \cite{ji2021improving} & - & - & - & - & - & - & 81.5 & 39.5 & 29.3 & 58.9 & 131.6 & 22.8 \\
 DLCT \cite{luo2021dual} & - & - & - & - & - & - & 81.4 & 39.8 & 29.5 & 59.1 & 133.8 & 23.0 \\
 RSTNet \cite{zhang2021rstnet} & - & - & - & - & - & - & 81.8 & 40.1 & 29.8 & 59.5 & 135.6 & 23.3 \\
 PureT \cite{wang2022end} & - & - & - & - & - & - & 82.1 & 40.9 & 30.2 & 60.1 & 138.2 & 24.2 \\
\hline 
 ExpansionNet v2 & 78.1 & 38.1 & 30.1 & 58.9 & 128.2 & 23.5 & \textbf{82.8} & \textbf{41.5} & \textbf{30.3} & \textbf{60.5} & \textbf{140.4} & \textbf{24.5} \\
 \hline
 \end{tabular}
\end{center}
\end{table*} 

\begin{table*}[htbp]
  \begin{center}
  \caption{Offline comparison of State-of-the-Art ensemble models over the Karpathy test split. B=BLEU. M=Meteor. R=Rouge. C=CIDEr-D. S=SPICE.}
\label{tab:offline_table_eval_ensemble}
  \begin{tabular}{| c | c  c  c  c  c  c | c  c  c  c  c  c |}
 \hline 
 {} & \multicolumn{6}{c|}{\textbf{Cross-Entropy}} & \multicolumn{6}{c|}{\textbf{CIDEr-D optimization}} \\
 \cline{2-13}
 \textbf{Model} & \textbf{B1} & \textbf{B4} & \textbf{M} & \textbf{R} & \textbf{C} & \textbf{S} & \textbf{B1} & \textbf{B4} & \textbf{M} & \textbf{R} & \textbf{C} & \textbf{S} \\
 \hline
 GCN-LSTM \cite{yao2018exploring} & 77.4 & 37.1 & 28.1 & 57.2 & 117.1 & 21.1 & 80.9 & 38.3 & 28.6 & 58.5 & 128.7 & 22.1\\
 SGAE \cite{yang2019auto} & - & - & - & - & - & - & 81.0 & 39.0 & 28.4 & 58.9 & 129.1 & 22.2\\
 AoANet \cite{huang2019attention} & 78.7 & 38.1 & 28.5 & 58.2 & 122.7 & 21.7 & 81.6 & 40.2 & 29.3 & 59.4 & 132.0 & 22.8\\
 X-Transformer \cite{pan2020x} & 77.8 & 37.7 & 29.0 & 58.0 & 122.1 & 21.9 & 81.7 & 40.7 & 29.9 & 59.7 & 135.3 & 23.8 \\
 GET \cite{ji2021improving} & - & - & - & - & - & - & 82.1 & 40.6 & 29.8 & 59.6 & 135.1 & 23.8 \\
 DLCT \cite{luo2021dual} & - & - & - & - & - & - & 82.2 & 40.8 & 29.9 & 59.8 & 137.5 & 23.3 \\
 PureT \cite{wang2022end} & - & - & - & - & - & - & 83.4 & 42.1 & 30.4 & 60.8 & 141.0 & 24.3 \\
 \hline
 ExpansionNet v2 & 78.5 & 38.5 & 29.9 & 58.8 & 128.7 & 23.6 & \textbf{83.5} & \textbf{42.7} & \textbf{30.6} & \textbf{61.1} & \textbf{143.7} & \textbf{24.7} \\
 \hline
 \end{tabular}
\end{center}
\end{table*}

Despite its apparent complexity, it is much more computationally friendly than the standard method consisting of a small batch size of 10 for 30 epochs for both optimization steps. As a matter of fact, only a much smaller number of training epochs are dedicated to fine-tuning the whole system. 
Thus, the time required for the calculation of the 
backbone's gradient is often avoided and the time required for forward operations can be drastically reduced as well. In particular, in our implementation, during steps 1 and 3 the backbone's forward pass is performed only once for each image in the data set. Therefore, its cost is replaced by a memory read and copy. All steps are trained using the RAdam optimizer \cite{liu2019variance} ($\beta_1=0.9, \ \beta_2=0.98$).

\subsection{Ablation Study}
\label{section_ablation}
To study the effectiveness of our method we replace the encoder and decoder in the baseline with our methods and evaluate several settings of expansion coefficients. It can be observed from Table \ref{tab:ablation_1} that the impact of the static expansion layer in the single group configuration is limited. In fact, it only slightly improves the baseline, regardless of the choice of $N_{E}$.
Conversely, the dynamic expansion layer showcases a more significant improvement obtaining the best result for $N_{E}=16$.
When the two expansion methods are combined, the model outperforms the baseline across all metrics with a margin of at least 
6.0 CIDEr-D, 2.0 BLEU, 0.5 SPICE, 1.0 ROUGE, and 0.8 METEOR. Analyzing several configurations of length groups in the static expansion, it appears that introducing more expansion vectors does not necessarily lead to better performances, since for $G=\{$128, 128, 128, 128, 128$\}$, $G=\{$384, 384, 384, 384, 384$\}$ and $G=\{$512, 512, 512, 512, 512$\}$ the model yield similar results. However, the model seems to benefit from a diverse selection of coefficients such as in the case of $G=\{$32, 64, 128, 256, 512$\}$ which will be adopted in the remaining experiments. Ultimately, all instances outperform the baseline across all metrics. 

\begin{table*}[htbp]
  \begin{center}
  
 \caption{Online server results on the MS-COCO 2014 test set which ground truth is unknown. B=BLEU. M=METEOR. R=ROUGE. C=CIDEr-D. S=SPICE.}
 \label{tab:online_table_eval}
  \begin{tabular}{ | c | c  c |  c  c |  c  c | c  c | c  c | c  c | c  c |}
 \hline
 {} & \multicolumn{2}{c|}{\textbf{B1}} & \multicolumn{2}{c|}{\textbf{B2}} &\multicolumn{2}{c|}{\textbf{B3}} & \multicolumn{2}{c|}{\textbf{B4}} & \multicolumn{2}{c|}{\textbf{METEOR}} & \multicolumn{2}{c|}{\textbf{ROUGE-L}} & \multicolumn{2}{c|}{\textbf{CIDEr-D}} \\
 \cline{2-15}
 {\textbf{Model}} & \textbf{c5} & \textbf{c40} & \textbf{c5} & \textbf{c40} & \textbf{c5} & \textbf{c40} & \textbf{c5} & \textbf{c40} & \textbf{c5} & \textbf{c40} & \textbf{c5} & \textbf{c40} & \textbf{c5} & \textbf{c40} \\
 \hline
 % SCST \cite{rennie2017self} & 78.1 & 93.7 & 61.9 & 86.0 & 47.0 & 75.9 & 35.2 & 64.5 & 27.0 & 35.5 & 56.3 & 70.7 & 114.7 & 116.0 \\
 Up-Down \cite{anderson2018bottom}   & 80.2 & 95.2 & 64.1 & 88.8 & 49.1 & 79.4 & 36.9 & 68.5 & 27.6 & 36.7 & 57.1 & 72.4 & 117.9 & 120.5\\
 GCN-LSTM \cite{yao2018exploring} & - & - & 
 65.5 & 89.3 & 50.8 & 80.3 & 38.7 & 69.7 & 28.5 & 37.6 & 58.5 &
 73.4 & 125.3 & 126.5 \\
 SGAE \cite{yang2019auto}& 81.0 & 95.3 &
 65.6 & 89.5 & 50.7 & 80.4 & 38.5 & 69.7 & 28.2 & 37.2 & 58.6 & 73.6 & 123.8 & 126.5 \\
 AoANet \cite{huang2019attention}& 81.0 & 95.0 &
 65.8 & 89.6 & 51.4 & 81.3 & 39.4 & 71.2 & 29.1 & 38.5 & 58.9 & 74.5 & 126.9 & 129.6 \\
 X-Transformer \cite{pan2020x} & 81.9 & 95.7 & 66.9 & 90.5 & 52.4 & 82.5 & 40.3 & 72.4 & 29.6 & 39.2 & 59.5 & 75.0 & 131.1 & 133.5 \\
 RSTNet \cite{zhang2021rstnet} & 82.1 & 96.4 & 67.0 & 91.3 & 52.2 & 83.0 & 40.0 & 73.1 & 29.6 & 39.1 & 59.5 & 74.6 & 131.9 & 134.0 \\
 GET \cite{ji2021improving} & 81.6 & 96.1 & 66.5 & 90.9 & 51.9 & 82.8 & 39.7 & 72.9 & 29.4 & 38.8 & 59.1 & 74.4 & 130.3 & 132.5 \\
 DLCT \cite{luo2021dual} & 82.4 & 96.6 & 67.4 & 91.7 & 52.8 & 83.8 & 40.6 & 74.0 & 29.8 & 39.6 & 59.8 & 75.3 & 133.3 & 135.4 \\
 PureT \cite{wang2022end} & 82.8 & 96.5 & 68.1 & 91.8 & 53.6 & 83.9 & 41.4 & 74.1 & 30.1 & 39.9 & 60.4 & 75.9 & 136.0 & 138.3 \\
 OFA \cite{wang2022ofa} & \textbf{84.5} & \textbf{98.1} & 70.1 & \textbf{94.4} & \textbf{55.9} & \textbf{87.8} & \textbf{43.6} & \textbf{78.7} & \textbf{32.1} & \textbf{42.7} & \textbf{62.5} & \textbf{79.0} & \textbf{147.2} & 149.6 \\
GIT \cite{wang2022git} & 84.3 & \textbf{98.1}  & 70.0 & \textbf{94.4} & 55.7 & 87.6 & 43.2 & 78.3 & 31.9 & 42.1 & 62.0 & 78.4 & 146.4 & \textbf{149.8} \\
\hline
ExpansionNet v2 & 83.3 & 96.9 & 8.8 & 92.6 & 54.4 & 85.0 & 42.1 & 75.3 & 30.4 & 40.1 & 60.8 & 76.4 & 138.5 & 140.8 \\
\hline
 \end{tabular}
 \end{center}
\end{table*} 

\subsection{Performance Comparison}
\label{sec:perfomance_comparision}

\subsubsection{COCO Offline Evaluation} Table \ref{tab:offline_table_eval} and Table \ref{tab:offline_table_eval_ensemble} report the score comparison between ExpansionNet v2 and the best-performing models in recent years.
Up-Down \cite{anderson2018bottom} introduced the idea of extracting a collection of features from the images using an object detector like Faster-RCNN \cite{ren2015faster} in contrast to the classification backbone \cite{xu2015show}. The idea was adopted in most of the following architectures as well, for instance, in the case of GCN-LSTM \cite{yao2018exploring} and SGAE \cite{yang2019auto}, which additionally implemented a convolutional graph network on top of it to exploit the information provided by a scene graph. AoANet \cite{huang2019attention} adopted the Transformer and improved the attentive components with two gates serving the purpose of simulating an additional level of attention over the inputs and augmented the language modelling part with an LSTM. On the other hand, X-Transformer \cite{pan2020x} adopted a fully attentive architecture and further refined the attentive blocks by means of bilinear pooling techniques. The most recent and performing architectures focused on increasing the more effective ways to feed visual information into the sequence modelling network. For instance, RSTNet \cite{zhang2021rstnet} showcased the effectiveness of grid features over regions, GET \cite{ji2021improving} processed the images using a global representation in conjunction with the local ones, DLCT \cite{luo2021dual} instead exploited the advantages of both regions and grid visual features.
Finally, PureT \cite{wang2022end} implemented the first end-to-end Transformer architecture applying the Window / Shifted-Window MHA \cite{liu2021swin} in both the encoder and decoder.
ExpansionNet v2 outperforms PureT by a margin of 0.7 BLEU1, 0.6 BLEU4, 0.1 METEOR, 0.4 ROUGE, 2.2 CIDEr-D and 0.3 SPICE in the single model case and by 0.1 BLEU1, 0.6 BLEU4, 0.3 ROUGE, 2.7 CIDEr-D and 0.4 SPICE in the ensemble configuration.

\subsubsection{COCO Online Evaluation} We evaluate ExpansionNet v2 using the ensemble configuration 
and adopting the standard Beam Search (beam size 5) over the official testing set of 40775 images, 
submitting the predictions to the online testing server. Results are reported in Table \ref{tab:online_table_eval}. c5 and c40 represent the scores of 5 and 40 reference captions (unknown to the user), respectively. 

Our model achieves State-of-the-Art performance (as of 2 July 2022) among non-generative models trained on MS-COCO 2014, outperforming the previous one \cite{wang2022end} by a margin of 1.2 BLEU4 (c40), 0.2 METEOR (c40), 0.5 ROUGE-L (c40) and 2.5 CIDEr-D in both c5 and c40 instances. However, it is ultimately outperformed, by a significant margin, by generative models \cite{wang2022ofa, wang2022git} which we consider orthogonal to our work since they focus more on training method and data quality rather than architecture design.

\subsubsection{Nocaps Evaluation} 
We evaluate ExpansionNet v2 over the nocaps validation set. In particular, we adopt a single model trained exclusively on Cross-Entropy Loss, using no
additional pre-training data sets. The predictions are generated by the standard Beam Search algorithm (beam size 3) in contrast to the CBS \cite{anderson2016guided}. A limited comparison is reported 
in Table \ref{tab:no_caps_eval}, which showcases that our model achieves very competitive results among the architectures trained in similar 
configurations, with an overall lead of 17.6 CIDEr and 1.4 SPICE over the Up-Down model \cite{anderson2018bottom}. 
It is still ultimately outperformed by recent V+L pre-training-based works 
\begin{table}[htbp]
\begin{center}
 \caption{Performances on nocaps validation set. C and S denote the CIDEr-D and SPICE scores respectively.}
  \label{tab:no_caps_eval}
  \begin{tabular}{| c | c | c | c | c |}
 \hline
\textbf{Domain} & \textbf{Metric} & \textbf{Enc-Dec}\cite{changpinyo2021conceptual} & \textbf{Up-Down}\cite{anderson2018bottom} & \textbf{Ours} \\
 \hline
 \multirow{2}{*}{In} & C & 72.8 & 78.1 & \textbf{83.8} \\
  & S & 11.1 & 11.6 & \textbf{12.6} \\
 \hline
 \multirow{2}{*}{Near} & C & 57.1 & 57.7 & \textbf{79.2} \\
 & S & 10.2 & 10.3 & \textbf{12.4} \\
 \hline
 \multirow{2}{*}{Out} & C & 34.1 & 31.3 & \textbf{54.0} \\
 & S & 8.3 & 8.3 & \textbf{9.3} \\
 \hline
 \multirow{2}{*}{All} & C & 54.7 & 55.3 & \textbf{72.9} \\  
 & S & 10.0 & 10.1 & \textbf{11.4} \\
 \hline
 \end{tabular}
\end{center}
\end{table}
such as \cite{hu2022scaling, zhang2021vinvl, li2022blip}.

\subsection{Training and Inference Cost}

The efficiency aspect was addressed in the design of the Expansion mechanism. For instance, it can be observed from Table \ref{tab:flops_ops} that doubling the expansion coefficients does not lead to double FLOPS, which would be the case of actually doubling the input sequence length. In particular, for small parameters, our model is comparable to the Transformer in terms of computational cost. In contrast, ExpansionNet v2 is 1.63$\times$ slower than the baseline because of an abundant selection of expansion coefficients.

In Table \ref{tab:training_time}, we compare our training time with the ones presented by other works. In particular, 
time entries are estimated assuming all model computational costs are the same as the ExpansionNet v2,  which is a generous approximation compared to generative models whose sizes are tens of times larger. 
Despite such premise and the fact that we also perform end-to-end training, it can be seen that our model can be trained up to 2.8$\times$ faster than other non-generative models and up to 46.8$\times$ faster in the case of generative ones. Recalling the results in Table \ref{tab:online_table_eval}, performance-wise, our model achieves 93.9\% performances of the State-of-the-Art model GIT \cite{wang2022git} but uses 7080$\times$ less data and is 129$\times$ smaller. 
\begin{table}[htbp]
  \begin{center}
 \caption{Inference cost comparison of ablation models on the MS-COCO 2014 validation set (5000 images).}
 \label{tab:flops_ops}
  \begin{tabular}{ | c | c | c |} 
\hline
 \textbf{Encoder} & \textbf{Decoder} & \textbf{FLOPS} \\
 \hline
 Baseline & Baseline & $9.28\times10^{12}$  \\
 Stc. Exp. G=$\{ 16 \}$ & Baseline & $9.62\times10^{12}$ \\
 Stc. Exp. G=$\{ 32 \}$ & Baseline & $9.70\times10^{12}$ \\
 Stc. Exp. G=$\{ 64 \}$ & Baseline & $9.88\times10^{12}$ \\
Baseline & Dyn. Exp.\textsubscript{$N_E$=4} & $9.40\times10^{12}$ \\

Baseline & Dyn. Exp.\textsubscript{$N_E$=8} & $9.43\times10^{12}$ \\
Baseline & Dyn. Exp.\textsubscript{$N_E$=16} & $9.48\times10^{12}$ \\

Stc. Exp. G=$\{ 64 \}$ & Dyn. Exp.\textsubscript{$N_E$=16} & $10.08\times10^{12}$ \\
Stc. Exp. G=$\{128\}$$\times5$ & Dyn. Exp.\textsubscript{$N_E$=16} & $13.26\times10^{12}$ \\
Stc. Exp. G=$\{256\}$$\times5$ & Dyn. Exp.\textsubscript{$N_E$=16} & $16.80\times10^{12}$ \\
Stc. Exp. G=$\{512\}$$\times5$ & Dyn. Exp.\textsubscript{$N_E$=16} & $23.88\times10^{12}$ \\
\hline
 ExpansionNet v2 & ExpansionNet v2 & $15.21\times10^{12}$ \\
 \hline
 \end{tabular}
 \end{center}
\end{table}

\begin{table*}[htbp]
  \begin{center}
   \caption{Training time comparison of State-of-the-Art works against our solution. ``Time'' represents the estimated time required to train models on a single NVIDIA A100 using the described strategy. $\gamma$, $\theta$ and $\sigma$ denote the normalized quantity of the number of parameters, the number of training images and the training cost compared to our proposal. The "$\star$" symbol denotes generative modes, typically pre-trained on multiple tasks and images from various sources. We simplify the matter using the cost of Cross-Entropy training on MS-COCO 2014, and the downstream task learning cost is ignored since it is negligible compared to the pre-training phase. }
\label{tab:training_time}
  \begin{tabular}{ | >{\RaggedRight\arraybackslash}p{2.5cm} | >{\RaggedRight\arraybackslash}p{2.0cm} | >{\RaggedRight\arraybackslash}p{5.0cm} | >{\RaggedRight\arraybackslash}p{4.5cm} | c  | }
 \hline
 \textbf{Source} & \textbf{Params. ($\gamma$)}  &  \textbf{Datasets $\rightarrow$ total num. images ($\theta$)} & \textbf{Training Description} & \textbf{Train. time ($\sigma$)} \\
 \hline
 Obj. Transf. \cite{herdade2019image}, \newline AoANet \cite{huang2019attention} \newline
 PureT \cite{wang2022end} & 33M (0.86) \newline 87M (2.28) \newline 34M (0.89)
  & 
 MS-COCO 2014 $\rightarrow$ 113K (1.00) &
 Cross-Entropy: $\sim$ 30 epochs and batch size 10.
 Reinforcement: 30 epochs and batch size 10. & 7 days (2.80)  \\
 X-Transformer \cite{pan2020x} & 141M (3.71) &  MS-COCO 2014 $\rightarrow$ 113K (1.00) & 
 Cross-Entropy: 70 epochs and batch size 40. 
 Reinforcement: 35 epochs and batch size 32. & 5 days (2.00) \\
 GIT \cite{wang2022git} $\star$ & 4.9B (128.94) &
 MS-COCO, CC3M, CC12M, VG, SBU, ALT200M + 0.6B $\rightarrow$ 0.8B (7079.64).  
 &
 % we considered only captioning images in this. GIT reports  54M instead of 15 for OFA, for this reason 
 2 epochs and we assume batch size 48 in the estimation. & 117 days (46.80) \\
 OFA \cite{wang2022ofa} $\star$ & 871M (22.92) &
 MS-COCO, CC3M, CC12M, VG, \newline SBU $\rightarrow$ 15M (132.74). 
 & 
 40 epochs and assume batch size 48 in the estimation. & 44 days (17.60) \\
 \hline
{ExpansionNet v2} & 38M (1.00) & MS-COCO 2014 $\rightarrow$ 113K (1.00) &  See Section \ref{training_section}. 
& 2.5 days (1.00) \\
\hline
 \end{tabular}
\end{center}
\end{table*}

\begin{table}[htbp]
 \begin{center}
 \caption{Examples of captions.}
\label{examples_1}
 \scriptsize
  \begin{tabular}[h]{| @{}c@{} | >{\RaggedRight\arraybackslash}p{4.0cm} |}
 \hline
 \textbf{Image} &\textbf{Captions} \\
 \hline
 % 1
 \includegraphics[width=0.18\textwidth, height=0.16\textwidth,  valign=m]{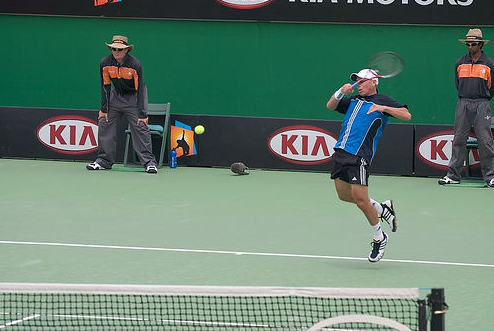}
& \vspace{-1.6cm}{
\leavevmode\newline
\textbf{Baseline:} A man holding a tennis ball on a tennis court. \leavevmode\newline
\textbf{ExpansionNet v2:} A man jumping in the air to hit a tennis ball. \leavevmode\newline
\textbf{Gt:} \{A tennis player jumps and swats at the ball.; A tennis player hitting a tennis ball on a court.; Professional tennis player immediately after returning a shot.\} } \\ 
\hline
 \includegraphics[width=0.18\textwidth, height=0.16\textwidth,  valign=m]{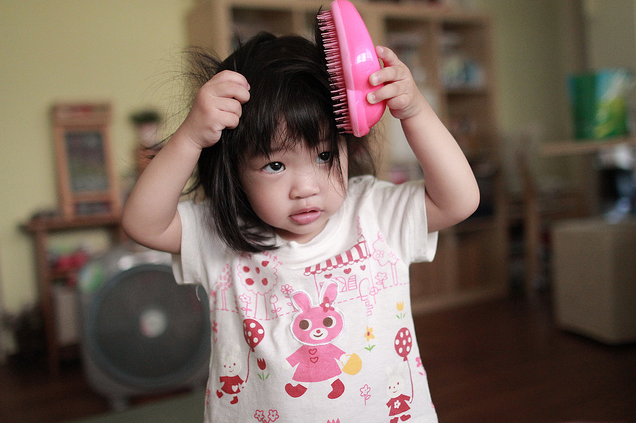}
& \vspace{-1.6cm}{ 
\leavevmode\newline
\textbf{Baseline:} A little girl brushing her hair with a table. \leavevmode\newline
\textbf{ExpansionNet v2:} A little girl brushing her hair with a pink brush.
\leavevmode\newline
 \textbf{Gt:} \{A young girl tries to comb her own hair.; A young child brushing her hair with a big pink brush.; A young girl is trying to brush her hair with a pink brush.\} } \\ 
 \hline
% \includegraphics[width=0.16\textwidth, height=0.15\textwidth,  valign=m]{img/img_example_4.png}
%& \vspace{-1.6cm}{
%\leavevmode\newline
%\textbf{Baseline:} Three horses grazing in the snow with a sheep. \leavevmode\newline
%\textbf{ExpansionNet v2:} Three horses grazing in the snow in a field.
% \leavevmode\newline \textbf{Gt:} \{Two brown horses standing next to a white horse on a snow covered field.; A group of horses grazing in the snow.; Horses eating grass through the snow in a field.\} } \\
% \hline
\end{tabular}
\end{center}
\end{table}

\subsection{Qualitative Analysis}

\begin{figure}[htbp]
\begin{center}
\includegraphics[width=0.3\textwidth]{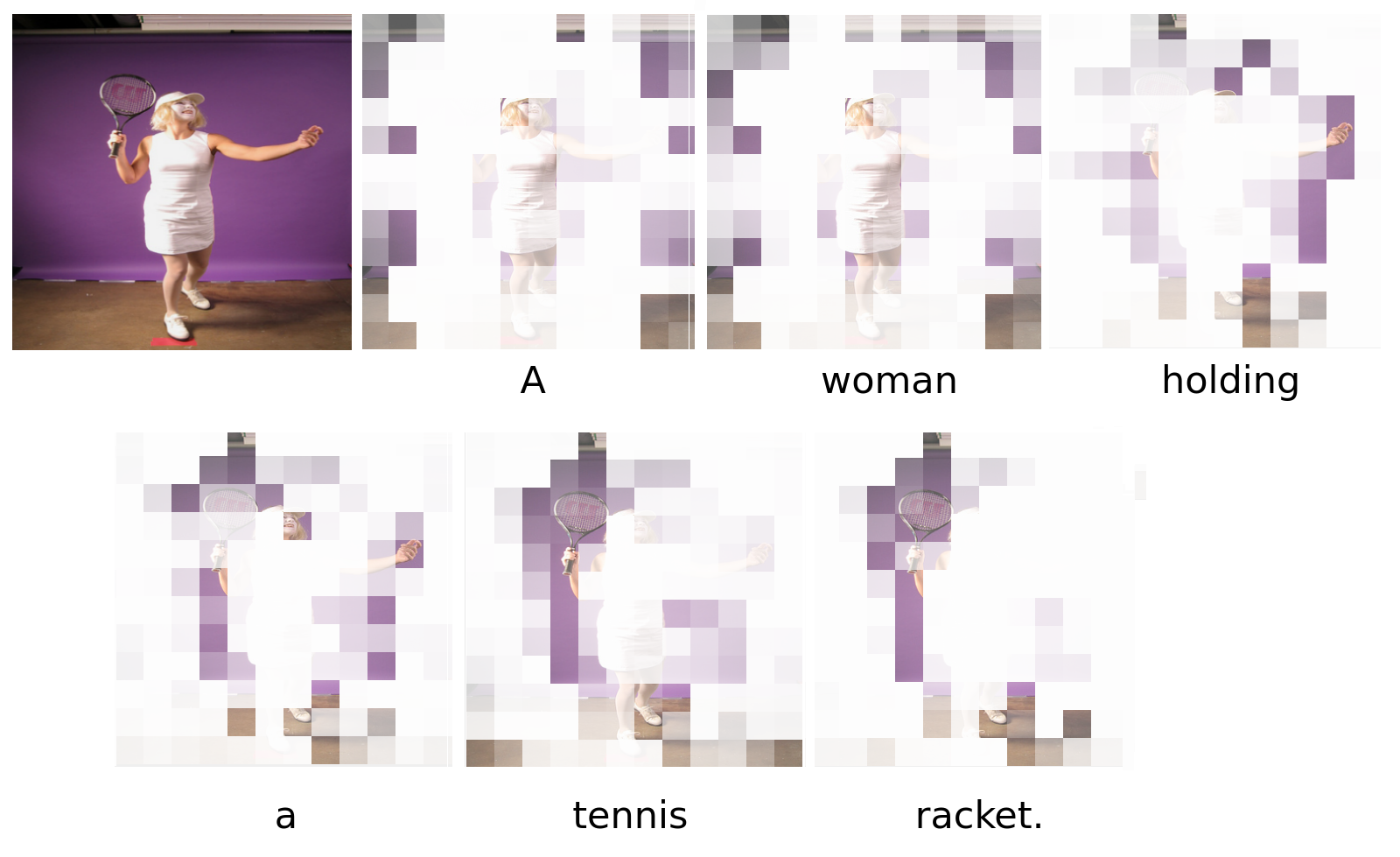}
\caption{Attention visualization of a single decoder head in ExpansionNet v2.}
\label{fig:attention_vis}
\end{center}
\end{figure}

Table \ref{examples_1} provides some examples of captions. Regardless of the image complexity, ExpansionNet v2 is not only able to correctly describe the subjects depicted in the scenes but also showcases a good level of semantic understanding by describing the goals and interactions. Unfortunately, our model seems to struggle with out-of-domain objects as showcased in Table \ref{nocaps_img} where, due to objects and terms unknown to the model, predictions are either imprecise (2\textsuperscript{nd} image) or incorrect (1\textsuperscript{st} image). Nonetheless, it appears to provide a roughly correct description of the image.  We showcase an example of attention visualization in Fig. \ref{fig:attention_vis}, where the scattered focus correctly outlines the main subjects despite the absence of an object detector.

\section{Conclusion}

In this work, we addressed the question of whether the fixed number of elements of the inputs represented a performance bottleneck in modern image-captioning systems. To this end, we presented the idea of an Expansion mechanism
and provided two concrete implementations called Static Expansion and Dynamic Expansion, that process the input using sequences that feature a different length compared to the one provided in the input. 
Upon these layers, we designed a new architecture called ExpansionNet v2 and trained it on the MS-COCO 2014 dataset using a fast End to End training approach. Extensive experiments conducted on the testing set showcase that our method  achieved better performances when compared to the baseline. This answer positively the initial research question of whether the input length can represent a bottleneck to the sequence processing. Additionally, ExpansionNet v2 achieved strong performances on both offline (143.7 CIDEr-D) and online (140.8 CIDEr-D) test splits and is outperformed mainly by V+L 
pre-training models, which we consider orthogonal to our work due to the differences in model size and additional training data.
In conclusion, we introduced the Expansion layers and ExpansionNet v2 and found the answer to our research question in the case of the Image Captioning field. Future works will further develop the methods and ideas presented in this work, motivated by the fact that they can be easily integrated into other solution approaches (such as V+L pre-training) and other research fields.

% - - - NoCaps Examples
\begin{table}[htbp]
 \begin{center}
 \scriptsize
 \caption{Examples on nocaps out-of-domain images.}
\label{nocaps_img}
 \begin{tabular}{ | @{}c@{} | >{\RaggedRight\arraybackslash}p{4.5cm} | }
 \hline
 \textbf{Image} & \textbf{Captions} \\
 \hline
 \includegraphics[height=0.14\textwidth, width=0.17\textwidth,  valign=m]{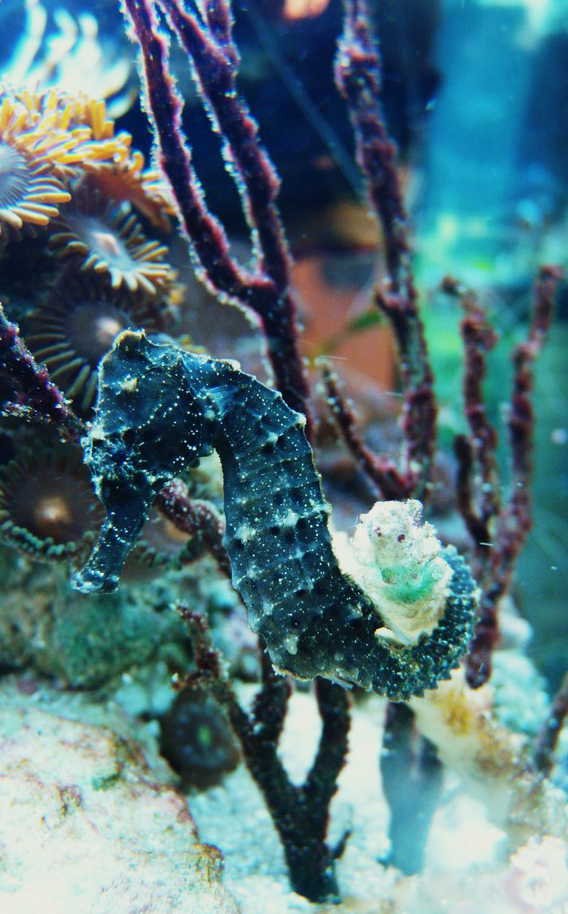}
& \vspace{-1.2cm}{\textbf{ExpansionNet v2:} A close-up of a fish in a body of water.
\leavevmode\newline
\textbf{3 gts:} \{ A seahorse in an aquarium full of water with some plants growing in the background.; A blue seahorse is swimming near sea plants on back.; A very small seahorse is in the water along with other pieces. \} } \\
\hline
\includegraphics[height=0.14\textwidth, width=0.17\textwidth, valign=m]{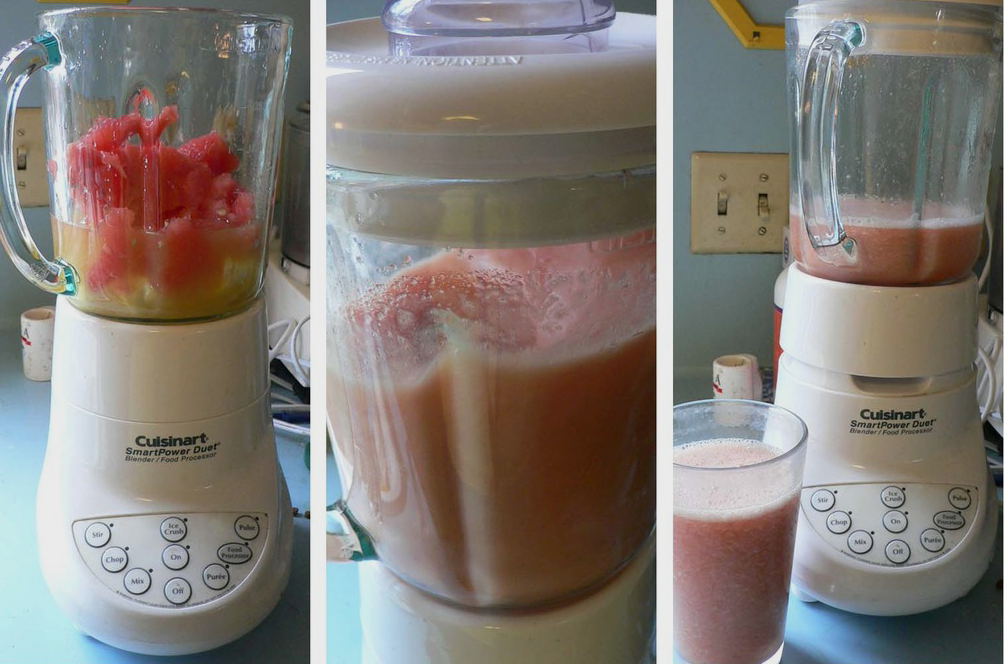}
& \vspace{-1.2cm}{\textbf{ExpansionNet v2:} Three pictures of a blender with red liquid in it.
\leavevmode\newline
\textbf{3 gts:} \{ A picture of three blenders with a strawberry looking beverage inside.; A white mixer in the process of making a smoothie.; The steps of making a smoothie in a blender are shown. \} } \\
\hline
\end{tabular}
\end{center}
\end{table}

% head viualization

\bibliographystyle{IEEEtran}
\bibliography{x_bibliography}

\begin{thebibliography}{10}
\providecommand{\url}[1]{#1}
\csname url@samestyle\endcsname
\providecommand{\newblock}{\relax}
\providecommand{\bibinfo}[2]{#2}
\providecommand{\BIBentrySTDinterwordspacing}{\spaceskip=0pt\relax}
\providecommand{\BIBentryALTinterwordstretchfactor}{4}
\providecommand{\BIBentryALTinterwordspacing}{\spaceskip=\fontdimen2\font plus
\BIBentryALTinterwordstretchfactor\fontdimen3\font minus \fontdimen4\font\relax}
\providecommand{\BIBforeignlanguage}[2]{{%
\expandafter\ifx\csname l@#1\endcsname\relax
\typeout{** WARNING: IEEEtran.bst: No hyphenation pattern has been}%
\typeout{** loaded for the language `#1'. Using the pattern for}%
\typeout{** the default language instead.}%
\else
\language=\csname l@#1\endcsname
\fi
#2}}
\providecommand{\BIBdecl}{\relax}
\BIBdecl

\bibitem{mitchell2012midge}
M.~Mitchell \emph{et~al.}, ``Midge: Generating image descriptions from computer vision detections,'' in \emph{Proceedings of the 13th Conference of the European Chapter of the Association for Computational Linguistics}, 2012, pp. 747--756.

\bibitem{kulkarni2013babytalk}
G.~Kulkarni \emph{et~al.}, ``Babytalk: Understanding and generating simple image descriptions,'' \emph{IEEE Transactions on Pattern Analysis and Machine Intelligence}, vol.~35, no.~12, pp. 2891--2903, 2013.

\bibitem{vinyals2015show}
O.~Vinyals, A.~Toshev, S.~Bengio, and D.~Erhan, ``Show and tell: A neural image caption generator,'' in \emph{Proceedings of the IEEE conference on computer vision and pattern recognition}, 2015, pp. 3156--3164.

\bibitem{xu2015show}
K.~Xu \emph{et~al.}, ``Show, attend and tell: Neural image caption generation with visual attention,'' in \emph{International conference on machine learning}, 2015, pp. 2048--2057.

\bibitem{anderson2018bottom}
P.~Anderson \emph{et~al.}, ``Bottom-up and top-down attention for image captioning and visual question answering,'' in \emph{Proceedings of the IEEE conference on computer vision and pattern recognition}, 2018, pp. 6077--6086.

\bibitem{ren2015faster}
S.~Ren, K.~He, R.~Girshick, and J.~Sun, ``Faster r-cnn: Towards real-time object detection with region proposal networks,'' \emph{arXiv preprint arXiv:1506.01497}, 2015.

\bibitem{hochreiter1997long}
S.~Hochreiter and J.~Schmidhuber, ``Long short-term memory,'' \emph{Neural computation}, vol.~9, no.~8, pp. 1735--1780, 1997.

\bibitem{cho2014learning}
K.~Cho \emph{et~al.}, ``Learning phrase representations using rnn encoder-decoder for statistical machine translation,'' \emph{arXiv preprint arXiv:1406.1078}, 2014.

\bibitem{bahdanau2014neural}
D.~Bahdanau, K.~Cho, and Y.~Bengio, ``Neural machine translation by jointly learning to align and translate,'' \emph{arXiv preprint arXiv:1409.0473}, 2014.

\bibitem{vaswani2017attention}
A.~Vaswani \emph{et~al.}, ``Attention is all you need,'' in \emph{Advances in neural information processing systems}, 2017, pp. 5998--6008.

\bibitem{wang2022end}
Y.~Wang, J.~Xu, and Y.~Sun, ``End-to-end transformer based model for image captioning,'' \emph{arXiv preprint arXiv:2203.15350}, 2022.

\bibitem{nguyen2022grit}
V.-Q. Nguyen, M.~Suganuma, and T.~Okatani, ``Grit: Faster and better image captioning transformer using dual visual features,'' in \emph{Computer Vision--ECCV 2022: 17th European Conference, Tel Aviv, Israel, October 23--27, 2022, Proceedings, Part XXXVI}.\hskip 1em plus 0.5em minus 0.4em\relax Springer, 2022, pp. 167--184.

\bibitem{liu2021swin}
Z.~Liu \emph{et~al.}, ``Swin transformer: Hierarchical vision transformer using shifted windows,'' in \emph{Proceedings of the IEEE/CVF International Conference on Computer Vision}, 2021, pp. 10\,012--10\,022.

\bibitem{pan2020x}
Y.~Pan, T.~Yao, Y.~Li, and T.~Mei, ``X-linear attention networks for image captioning,'' in \emph{Proceedings of the IEEE/CVF Conference on Computer Vision and Pattern Recognition}, 2020, pp. 10\,971--10\,980.

\bibitem{wang2022ofa}
P.~Wang \emph{et~al.}, ``Ofa: Unifying architectures, tasks, and modalities through a simple sequence-to-sequence learning framework,'' in \emph{International Conference on Machine Learning}.\hskip 1em plus 0.5em minus 0.4em\relax PMLR, 2022, pp. 23\,318--23\,340.

\bibitem{wang2022git}
J.~Wang \emph{et~al.}, ``Git: A generative image-to-text transformer for vision and language,'' \emph{arXiv preprint arXiv:2205.14100}, 2022.

\bibitem{herdade2019image}
S.~Herdade, A.~Kappeler, K.~Boakye, and J.~Soares, ``Image captioning: Transforming objects into words,'' \emph{arXiv preprint arXiv:1906.05963}, 2019.

\bibitem{luo2021dual}
Y.~Luo \emph{et~al.}, ``Dual-level collaborative transformer for image captioning,'' in \emph{Proceedings of the AAAI Conference on Artificial Intelligence}, vol.~35, no.~3, 2021, pp. 2286--2293.

\bibitem{guo2019star}
Q.~Guo, X.~Qiu, P.~Liu, Y.~Shao, X.~Xue, and Z.~Zhang, ``Star-transformer,'' \emph{arXiv preprint arXiv:1902.09113}, 2019.

\bibitem{hao2019modeling}
J.~Hao, X.~Wang, B.~Yang, L.~Wang, J.~Zhang, and Z.~Tu, ``Modeling recurrence for transformer,'' \emph{arXiv preprint arXiv:1904.03092}, 2019.

\bibitem{luong2015effective}
M.-T. Luong, H.~Pham, and C.~D. Manning, ``Effective approaches to attention-based neural machine translation,'' \emph{arXiv preprint arXiv:1508.04025}, 2015.

\bibitem{raganato2020fixed}
A.~Raganato, Y.~Scherrer, and J.~Tiedemann, ``Fixed encoder self-attention patterns in transformer-based machine translation,'' \emph{arXiv preprint arXiv:2002.10260}, 2020.

\bibitem{tay2021synthesizer}
Y.~Tay, D.~Bahri, D.~Metzler, D.-C. Juan, Z.~Zhao, and C.~Zheng, ``Synthesizer: Rethinking self-attention for transformer models,'' in \emph{International conference on machine learning}.\hskip 1em plus 0.5em minus 0.4em\relax PMLR, 2021, pp. 10\,183--10\,192.

\bibitem{ramsauer2020hopfield}
H.~Ramsauer \emph{et~al.}, ``Hopfield networks is all you need,'' \emph{arXiv preprint arXiv:2008.02217}, 2020.

\bibitem{you2020hard}
W.~You, S.~Sun, and M.~Iyyer, ``Hard-coded gaussian attention for neural machine translation,'' \emph{arXiv preprint arXiv:2005.00742}, 2020.

\bibitem{lee2021fnet}
J.~Lee-Thorp, J.~Ainslie, I.~Eckstein, and S.~Ontanon, ``Fnet: Mixing tokens with fourier transforms,'' \emph{arXiv preprint arXiv:2105.03824}, 2021.

\bibitem{tolstikhin2021mlp}
I.~O. Tolstikhin \emph{et~al.}, ``Mlp-mixer: An all-mlp architecture for vision,'' \emph{Advances in neural information processing systems}, vol.~34, pp. 24\,261--24\,272, 2021.

\bibitem{socher2010connecting}
R.~Socher and L.~Fei-Fei, ``Connecting modalities: Semi-supervised segmentation and annotation of images using unaligned text corpora,'' in \emph{2010 IEEE Computer Society Conference on Computer Vision and Pattern Recognition}.\hskip 1em plus 0.5em minus 0.4em\relax IEEE, 2010, pp. 966--973.

\bibitem{yao2010i2t}
B.~Z. Yao, X.~Yang, L.~Lin, M.~W. Lee, and S.-C. Zhu, ``I2t: Image parsing to text description,'' \emph{Proceedings of the IEEE}, vol.~98, no.~8, pp. 1485--1508, 2010.

\bibitem{wang2020show}
L.~Wang, Z.~Bai, Y.~Zhang, and H.~Lu, ``Show, recall, and tell: Image captioning with recall mechanism,'' in \emph{Proceedings of the AAAI Conference on Artificial Intelligence}, vol.~34, no.~07, 2020, pp. 12\,176--12\,183.

\bibitem{huang2019attention}
L.~Huang, W.~Wang, J.~Chen, and X.-Y. Wei, ``Attention on attention for image captioning,'' in \emph{Proceedings of the IEEE International Conference on Computer Vision}, 2019, pp. 4634--4643.

\bibitem{kim2018bilinear}
J.-H. Kim, J.~Jun, and B.-T. Zhang, ``Bilinear attention networks,'' \emph{arXiv preprint arXiv:1805.07932}, 2018.

\bibitem{sukhbaatar2019augmenting}
S.~Sukhbaatar, E.~Grave, G.~Lample, H.~Jegou, and A.~Joulin, ``Augmenting self-attention with persistent memory,'' \emph{arXiv preprint arXiv:1907.01470}, 2019.

\bibitem{gulati2020conformer}
A.~Gulati \emph{et~al.}, ``Conformer: Convolution-augmented transformer for speech recognition,'' \emph{arXiv preprint arXiv:2005.08100}, 2020.

\bibitem{ji2021improving}
J.~Ji \emph{et~al.}, ``Improving image captioning by leveraging intra-and inter-layer global representation in transformer network,'' in \emph{Proceedings of the AAAI conference on artificial intelligence}, vol.~35, no.~2, 2021, pp. 1655--1663.

\bibitem{cornia2020meshed}
M.~Cornia, M.~Stefanini, L.~Baraldi, and R.~Cucchiara, ``Meshed-memory transformer for image captioning,'' in \emph{Proceedings of the IEEE/CVF Conference on Computer Vision and Pattern Recognition}, 2020, pp. 10\,578--10\,587.

\bibitem{zeng2022s2}
P.~Zeng, H.~Zhang, J.~Song, and L.~Gao, ``S2 transformer for image captioning,'' in \emph{Proceedings of the International Joint Conferences on Artificial Intelligence}, vol.~5, 2022.

\bibitem{wang2022unifying}
P.~Wang \emph{et~al.}, ``Unifying architectures, tasks, and modalities through a simple sequence-to-sequence learning framework,'' \emph{arXiv preprint arXiv:2202.03052}, 2022.

\bibitem{hu2022scaling}
X.~Hu \emph{et~al.}, ``Scaling up vision-language pre-training for image captioning,'' in \emph{Proceedings of the IEEE/CVF Conference on Computer Vision and Pattern Recognition}, 2022, pp. 17\,980--17\,989.

\bibitem{li2022blip}
J.~Li, D.~Li, C.~Xiong, and S.~Hoi, ``Blip: Bootstrapping language-image pre-training for unified vision-language understanding and generation,'' \emph{arXiv preprint arXiv:2201.12086}, 2022.

\bibitem{lin2014microsoft}
T.-Y. Lin \emph{et~al.}, ``Microsoft coco: Common objects in context,'' in \emph{European conference on computer vision}.\hskip 1em plus 0.5em minus 0.4em\relax Springer, 2014, pp. 740--755.

\bibitem{xiong2020layer}
R.~Xiong \emph{et~al.}, ``On layer normalization in the transformer architecture,'' in \emph{International Conference on Machine Learning}.\hskip 1em plus 0.5em minus 0.4em\relax PMLR, 2020, pp. 10\,524--10\,533.

\bibitem{rennie2017self}
S.~J. Rennie, E.~Marcheret, Y.~Mroueh, J.~Ross, and V.~Goel, ``Self-critical sequence training for image captioning,'' in \emph{Proceedings of the IEEE Conference on Computer Vision and Pattern Recognition}, 2017, pp. 7008--7024.

\bibitem{luo2020better}
R.~Luo, ``A better variant of self-critical sequence training,'' \emph{arXiv preprint arXiv:2003.09971}, 2020.

\bibitem{karpathy2015deep}
A.~Karpathy and L.~Fei-Fei, ``Deep visual-semantic alignments for generating image descriptions,'' in \emph{Proceedings of the IEEE conference on computer vision and pattern recognition}, 2015, pp. 3128--3137.

\bibitem{agrawal2019nocaps}
H.~Agrawal \emph{et~al.}, ``Nocaps: Novel object captioning at scale,'' in \emph{Proceedings of the IEEE/CVF International Conference on Computer Vision}, 2019, pp. 8948--8957.

\bibitem{deng2009imagenet}
J.~Deng, W.~Dong, R.~Socher, L.-J. Li, K.~Li, and L.~Fei-Fei, ``Imagenet: A large-scale hierarchical image database,'' in \emph{2009 IEEE conference on computer vision and pattern recognition}.\hskip 1em plus 0.5em minus 0.4em\relax Ieee, 2009, pp. 248--255.

\bibitem{hu2022exploring}
J.~C. Hu, R.~Cavicchioli, and A.~Capotondi, ``Exploring the sequence length bottleneck in the transformer for image captioning,'' 2022.

\bibitem{Hu202339}
J.~Hu, R.~Cavicchioli, and A.~Capotondi, ``A request for  clarity over  the  end of  sequence token in  the  self-critical sequence training",'' \emph{Lecture Notes in Computer Science (including subseries Lecture Notes in Artificial Intelligence and Lecture Notes in Bioinformatics)}, vol. 14233 LNCS, p. 39 – 50, 2023.

\bibitem{yao2018exploring}
T.~Yao, Y.~Pan, Y.~Li, and T.~Mei, ``Exploring visual relationship for image captioning,'' in \emph{Proceedings of the European conference on computer vision (ECCV)}, 2018, pp. 684--699.

\bibitem{yang2019auto}
X.~Yang, K.~Tang, H.~Zhang, and J.~Cai, ``Auto-encoding scene graphs for image captioning,'' in \emph{Proceedings of the IEEE/CVF Conference on Computer Vision and Pattern Recognition}, 2019, pp. 10\,685--10\,694.

\bibitem{zhang2021rstnet}
X.~Zhang \emph{et~al.}, ``Rstnet: Captioning with adaptive attention on visual and non-visual words,'' in \emph{Proceedings of the IEEE/CVF conference on computer vision and pattern recognition}, 2021, pp. 15\,465--15\,474.

\bibitem{liu2019variance}
L.~Liu \emph{et~al.}, ``On the variance of the adaptive learning rate and beyond,'' \emph{arXiv preprint arXiv:1908.03265}, 2019.

\bibitem{anderson2016guided}
P.~Anderson, B.~Fernando, M.~Johnson, and S.~Gould, ``Guided open vocabulary image captioning with constrained beam search,'' \emph{arXiv preprint arXiv:1612.00576}, 2016.

\bibitem{changpinyo2021conceptual}
S.~Changpinyo, P.~Sharma, N.~Ding, and R.~Soricut, ``Conceptual 12m: Pushing web-scale image-text pre-training to recognize long-tail visual concepts,'' in \emph{Proceedings of the IEEE/CVF Conference on Computer Vision and Pattern Recognition}, 2021, pp. 3558--3568.

\bibitem{zhang2021vinvl}
P.~Zhang \emph{et~al.}, ``Vinvl: Revisiting visual representations in vision-language models,'' in \emph{Proceedings of the IEEE/CVF conference on computer vision and pattern recognition}, 2021, pp. 5579--5588.

\end{thebibliography}

\end{document}